\documentclass[times, review, 10pt]{elsarticle}
\usepackage{tabularx}   
\usepackage{multirow} 
\usepackage{xcolor}
\usepackage{subcaption}
\usepackage{makecell}

\usepackage{amssymb}
\usepackage{amsmath}
\usepackage[ruled,vlined]{algorithm2e} 
\usepackage[utf8]{inputenc}
\usepackage[T1]{fontenc}
\usepackage{tipa}  
\usepackage{amsthm}
\usepackage{url}
\usepackage{arydshln}
\usepackage{float} 
\journal{Pattern Recognition Journal}

\begin{document}

\begin{frontmatter}




\title{A Human-in-the-Loop Label Error Detection Framework Applied to Arabic-Script HTR Datasets}



\author[label1]{Sana Al-azzawi}
\author[label1]{Elisa Barney}
\author[label1]{György Kovács}
\author[label1]{Marcus Liwicki}

\affiliation[label1]{
    organization={Luleå University of Technology, 
    Department of Computer Science, Electrical and Space Engineering},
    city={Luleå},
    postcode={97187},
    country={Sweden}
}
\begin{abstract}
Despite recent advances, Handwritten Text Recognition (HTR) for Arabic-script languages still lags behind Latin-script HTR. Part of the problem is dataset quality. To help closing this gap, we propose a two-stage framework (CER-HV) for detecting label errors. Stage 1 (CER) is a Character-Error-Rate-based noise detector built on a Convolutional Recurrent Neural Network (CRNN) architecture. Stage 2 (HV) is the Human-In-The-Loop (HITL) Verification of noisy samples detected by the first stage. Applying the CER-HV framework on multiple Arabic-script datasets can identify samples with label errors including transcription, segmentation, orientation, and non-text content errors that can markedly affect HTR performance. These errors were identified by the first stage of the framework with up to 90\% (top-50) precision.

We also show that our CRNN achieves state-of-the-art performance across five of the six evaluated datasets, reaching 8.46\% Character Error Rate (CER) on KHATT (Arabic), 8.22\% on PHTI (Pashto), 10.59\% on Ajami, and 10.11\% on Muharaf (Arabic), all without any data cleaning. We establish a new baseline of 11.3\% CER on the PHTD (Persian) dataset. Applying CER-HV improves evaluation CER by up to 1.8 percentage points after dataset cleaning and retraining. Although our experiments focus on documents written in an Arabic-script language, the framework is general and can be applied to other text recognition datasets.

\end{abstract}






\end{frontmatter}

\section{Introduction}
Handwritten documents preserve cultural heritage and historical knowledge across generations. To support the preservation, search, and analysis of these documents, Handwritten Text Recognition (HTR) aims to automatically transcribe them to machine-readable text~\cite{SANCHEZ2019122}. Among the different HTR settings, line-level recognition has become a practical and widely adopted choice because it balances annotation effort with the ability to capture meaningful contextual information~\cite{li2025htr}. In this work, we focus on line-level HTR, where the goal is to predict the character sequence contained in a single text line. Compared with character- or word-level recognition, this setting is the current research frontier and better reflects real document conditions~\cite{garrido2025handwritten}.

Currently, HTR for Arabic-script languages, 
still lags behind Latin-script HTR~\cite{aljishi2024comparative}. This gap is often attributed to the characteristics of Arabic script (cursive with letters differing only by small diacritical marks, and shapes changing depending on their position within a word) and its datasets being fewer, smaller, and less diverse than that of Latin-scripts~\cite{salaheldin2025advancements}. To reduce this performance gap, research has mostly focused on architectural improvements, paying much less attention to data quality and noisy labels.  For example, HATFormer ~\cite{chan2024hatformer} adapts TrOCR to historical Arabic manuscripts, while Aljishi et al. ~\cite{aljishi2024comparative} compare multiple deep HTR architectures, including Transformer-based models, for Arabic script.  

Creating handwritten text data sets is time and labor intensive. Often semi-automated methods are used and if human oversight is inadequate, samples with errors (e.g., incorrect or incomplete transcriptions, segmentation problems, orientation mistakes, script-mismatched samples or non-textual elements) are included in the final datasets. These label errors are problematic as deep neural networks can memorize noisy labels during training, which harms generalization~\cite{song2022learning} and can distort model selection and benchmarking~\cite{northcutt2021pervasive}. Recent work has also emphasized the importance of reliable evaluation protocols and benchmark assessment in HTR systems~\cite{vidal2023end}.  

While the impact of noisy labels has been studied extensively in computer vision (especially for classification tasks \cite{song2022learning}), it has received almost no attention in text recognition.

Prior work on learning-dynamics-based noise detection shows that neural networks tend to learn clean samples first, then memorize harder or noisy ones \cite{arpit2017closer,song2020does}. 
One example is O2U-Net \cite{huang2019o2u}, which ranks samples based on training loss, but the method was designed for classification tasks that use cross-entropy loss. 
In Connectionist Temporal Classification (CTC) based HTR, the loss is computed by summing the probabilities of all valid alignments (marginalizing over all paths) \cite{graves2006connectionist}.

Per-sample loss is influenced by alignment uncertainty and sequence length, making it less reliable for identifying mislabeled samples. Consequently, loss-based ranking is less informative for  noise detection in HTR, motivating our use of Character Error Rate (CER) instead.

Building on this insight, we propose a framework based on two key design choices. Firstly, we use CER for the ranking of errors, to match the sequence-to-sequence (seq2seq) formulation of our task. Secondly, we introduce early stopping to mitigate the effects of alignment uncertainty on learning dynamics in CTC-based sequence recognition. This yields an effective and practical noise detector for text recognition tasks such as HTR.


Automated scoring alone can mistakenly flag valid but difficult samples as noisy. Prior work has shown that only about half of algorithmically flagged candidates correspond to to actual label errors~\cite{northcutt2021pervasive}. Given the relatively small size of typical HTR datasets (e.g., IAM  $\approx 10K $  lines~\cite{marti2002iam}, KHATT  $\approx 6K $~\cite{mahmoud2014khatt}), human inspection of high-CER samples is practical. We therefore introduce CER-based Ranking with Human Verification (CER-HV), as a two-stage framework. To our knowledge, no prior work has combined noise detection with human verification in seq2seq tasks such as HTR.

\begin{figure}[t]
\centering

\includegraphics[width=0.8\textwidth]{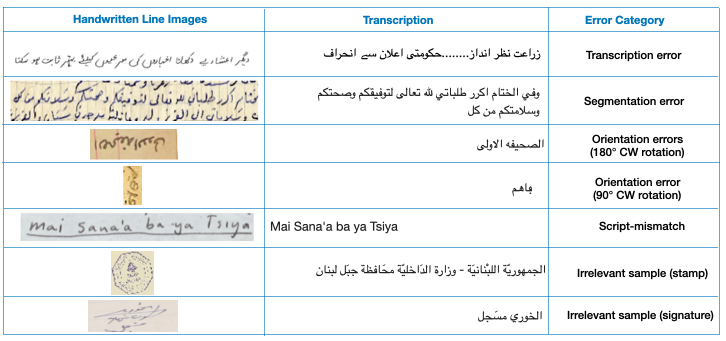}
\caption{Representative examples of error categories found in the datasets.}
\label{samples}
\end{figure}

Using this framework, we identified label errors across the training, validation, and test splits of five Arabic-script datasets. Although the overall proportion of errors is small in NUST-UHWR~\cite{ul2022convolutional}, KHATT~\cite{mahmoud2014khatt}, and PHTI~\cite{hussain2022phti}, higher concentrations are observed in the Muharaf~\cite{saeed2024muharaf} and Ajami~\cite{yousuf2025handwritten} datasets, where mislabeled samples constitute a non-negligible portion of the validation and test splits. Figure \ref{samples} presents  examples of these error types.

Overall, this study shows that data quality plays a major role in Arabic-script HTR and should be considered alongside model design. The proposed CER-HV framework helps identify and address label errors, improving the reliability of benchmark datasets in HTR. The main contributions of this work are summarized as follows:


\begin{enumerate}
\item We present the first systematic analysis of label and content errors in Arabic-script HTR datasets, defining a practical error taxonomy spanning transcription, segmentation, orientation, script mismatch, and non-text content.

\item We introduce CER-HV, a CER-based ranking with human verification framework for HTR that adapts learning-dynamics-based noise detection to CTC-based sequence recognition through CER-based scoring and early stopping. This can be used as a general procedure for inspecting and validating label quality in line-level HTR datasets.



\item We demonstrate that data quality can markedly distort reported HTR benchmarks, and quantify the impact of label noise across six Arabic-script datasets, showing evaluation CER reductions of up to 1.8 percentage points (pp) and establishing updated reference results on cleaned evaluation splits.

\item We establish a strong CRNN baseline for line-level Arabic-script HTR achieving state-of-the-art results on multiple datasets, including a reduction from 20.7\% to 8.22\% CER on Pashto (PHTI), and the first benchmarks for Ajami and Persian.

\item We provide the first publicly available cleaned evaluation splits and line-level benchmarks for Persian handwritten text (PHTD), including extracted line images and test partitions, enabling reproducibility and future benchmarking.

\item  All code, cleaned splits, human-verified error annotations, and pseudo-label candidates for transcription errors are released to support reproducible and reliable benchmarking in Arabic-script HTR.\footnote{\url{https://github.com/SanaNGU/CER-HV}}.
\end{enumerate}

The remainder of the paper is organized as follows. Section 2 reviews related work on Arabic-script HTR and learning with noisy labels. Section 3 describes the proposed two-stage framework, and explains how CER and learning dynamics are adapted to sequence prediction. Section 4 presents the experimental setup, including the data, preprocessing, and training details. Section 5 reports experimental results, and examines the effect of cleaning label errors on CER. Section 6 concludes the paper. 

\section{Related Work}

\subsection{Arabic-script HTR}
Arabic-script HTR remains challenging because the script is inherently cursive and context-dependent: the same letter may take different shapes depending on whether it appears in isolated, initial, medial, or final position, and many letters differ only by dots or optional diacritics (tashkīl/ḥarakāt), which are small and easily missed in degraded scans. These properties complicate segmentation and increase confusion among visually similar characters, especially in historical material with diverse calligraphic styles and layout noise \cite{saeed2024muharaf,salaheldin2025advancements, yousuf2025handwritten}.

Early Arabic-script HTR work utilized hidden markov models and handcrafted features, focusing primarily on isolated word or subword recognition within closed-set lexicons.
While these systems achieved high precision on fixed vocabularies, they relied on explicit or semi-explicit segmentation assumptions, such as predefined sliding windows, grapheme hypotheses, or lexicon-driven alignment. This reliance limited their robustness to unconstrained handwriting and hindered effective generalization to open-vocabulary, line-level transcription.

With the shift to deep learning, Arabic-script HTR increasingly moved toward segmentation-free sequence modeling that maps visual feature sequences directly to text, combining convolutional neural networks (CNNs) for feature extraction with recurrent sequence models (e.g., BLSTM, GRU, or MDLSTM) and training them using CTC \cite{graves2006connectionist}. A key motivation for adopting CTC  is that, in HTR the alignment between the input image sequence and the output character sequence is generally unknown; CTC enables end-to-end training by marginalizing over all valid alignments without requiring explicit frame-level or character-level segmentation. Within this paradigm, CRNN-style architectures based on a CNN–RNN–CTC design, as introduced by Shi et al. \cite{shi2016end}, became a common baseline, as they jointly learn visual representations and sequential dependencies while avoiding explicit segmentation. Variants of this design, often strengthened through deeper convolutional backbones, data augmentation, or architectural refinements, were subsequently adopted across multiple Arabic-script languages \cite{saeed2024muharaf,ul2022convolutional, bouchal2025towards}.

In line with developments in Latin-script HTR, Arabic-script research has increasingly explored Transformer-based architectures to replace or augment recurrence. By leveraging self-attention \cite{vaswani2017attention}, these models offer alternatives to CTC through seq2seq or transducer-style decoding \cite{momeni2024transformer}. For example, HATFormer adapted the English-centric TrOCR model to historical Arabic manuscripts, demonstrating that architectures built for Latin scripts can be transferred to Arabic with appropriate preprocessing and training strategies \cite{chan2024hatformer}.

A similar direction is observed in Urdu HTR, where Transformer-based architectures have been used \cite{riaz2022conv}, often relying on additional printed or synthetic text data to supplement handwritten training sets due to their data-intensive nature \cite{maqsood2023unified,hassan2025tda,hamza2024network}.

Another Arabic-script language that has received comparatively limited attention at the line level is Persian. Much of the existing work on Persian HTR focuses on isolated characters, digits, or word-level classification \cite{sadri2016novel}, largely due to the absence of publicly available line- or page-level Persian handwriting datasets. This motivates the use of the PHTD dataset \cite{alaei2012dataset} and the extraction of line images to enable systematic benchmarking of Persian HTR at the line level.

Progress on Pashto at the line level has long been constrained by limited data availability. Until recently, the lack of suitable datasets restricted most research to isolated characters or small-scale experiments. The introduction of the Pashto Handwritten Text Imagebase (PHTI) \cite{hussain2022phti} enabled the first large-scale benchmark on real-world Pashto handwritten text lines, where performance was evaluated using an MD-LSTM-based model \cite{hussain2024deep}. However, evaluation has largely remained confined to this single dataset, with limited comparative analysis across different recognition models.

The first publicly available Ajami HTR dataset was introduced only recently \cite{yousuf2025handwritten}, providing manually segmented and transcribed manuscripts for West African languages written in adapted variants of the Arabic script, specifically Hausa and Fulfulde. While this work demonstrates that existing Arabic-script recognition models perform poorly on Ajami manuscripts, it does not establish standardized benchmarks or comparative baselines. As a result, Ajami remains largely unexamined in systematic HTR evaluations, further motivating its inclusion in multi-language benchmarking studies.

A key gap across this literature is that Arabic-script HTR is still commonly pursued per language/script community in isolation (Arabic vs. Persian vs. Urdu vs. Ajami, etc.), despite substantial overlap across these languages. As a result, the extent to which a single, well-configured recognition model exhibits consistent behavior across different Arabic-script languages has not been systematically examined. We also show that such cross-language behavior can be assessed by benchmarking the same CRNN architecture across multiple Arabic-script datasets.

\subsection{Label Error Detection  in Handwritten Text Recognition}
Learning with noisy labels has been extensively studied in machine learning, primarily in the context of image classification. A comprehensive survey \cite{song2022learning} showed that most existing approaches are designed for fixed-label prediction problems trained with cross-entropy loss, and can be broadly grouped into loss-robust training, sample selection, label correction, and noise-transition modeling strategies. Within this literature, label error detection is typically achieved by exploiting training dynamics (e.g., small-loss behavior), prediction confidence, or agreement across models.

In contrast, label error detection has received comparatively little attention in HTR and related seq2seq text recognition tasks. Unlike classification, HTR models predict variable-length character sequences and are trained using sequence-level objectives, including alignment-based losses such as CTC as well as cross-entropy losses in encoder–decoder Transformer architectures \cite{garrido2025handwritten}. As a result, the training signal reflects alignment uncertainty and decoding effects in addition to transcription correctness, which makes the direct adoption of classification-oriented noise detection strategies less straightforward and limits the extent to which conclusions from the classification literature can be transferred to HTR without adaptation.

Several prominent noise detection approaches rely on model confidence estimates \cite{song2022learning}. Confident Learning, for example, identifies potential label errors by comparing predicted class probabilities against the given labels under a class-conditional noise assumption \cite{northcutt2021confident}. While such methods are effective for classification, their extension to sequence prediction tasks presents additional challenges, as confidence must be defined over entire sequences rather than individual classes and is often miscalibrated in practice. In particular, studies in Optical Character Recognition (OCR) report that word-level confidence scores do not consistently correlate with character or word error rates and depend strongly on  OCR system settings \cite{cuper2023unraveling}. Similarly, recent work on scene-text recognition demonstrates that modern sequence models are systematically overconfident, and that raw confidence scores do not reliably reflect true transcription accuracy \cite{slossberg2022calibration}. These findings indicate that confidence-based noise detection approaches, such as Confident Learning, remain challenging to apply to seq2seq text recognition tasks like HTR. For these reasons, this work focuses on CER-based ranking as a simple and task-aligned alternative for sequence recognition settings. Systematic comparison against Confident Learning and ensemble-based filtering strategies in seq2seq recognition remains an important direction for future investigation.

Another family of methods exploits learning dynamics to identify noisy labels. O2U-Net ranks samples based on their average training loss over cycles of underfitting and overfitting, under the observation that mislabeled samples tend to maintain higher loss throughout training \cite{huang2019o2u}. This idea is closely related to the memorization effect, where deep networks fit clean samples earlier than noisy ones. This early-learning phenomenon has been analyzed more broadly in the context of learning with noisy labels, where several works show that deep networks first fit clean labels before memorizing noisy ones, and propose training strategies that explicitly exploit or regularize this behavior to improve robustness 
\cite{ xia2020robust}.
While O2U-Net is formulated for classification, its reliance on per-sample loss trajectories makes it conceptually applicable to other settings. However, in HTR, the interpretation of per-sample CTC loss is complicated by alignment variability and structural errors, and cyclic overfitting strategies may not always be desirable. Prior work on early stopping shows that preventing late-stage memorization can be beneficial under label noise, suggesting that careful control of training dynamics is important when adapting such methods \cite{song2020does}.

Overall, the literature suggests that while label error detection is well studied for classification, there is limited work that directly addresses noisy supervision 
in seq2seq text recognition, including HTR, OCR, and scene-text recognition. This gap motivates strategies for identifying samples with errors in HTR, forming the basis of the approach explored in this work.

\section{Method}

\begin{figure}[t]
\centering
\includegraphics[width=1\textwidth]{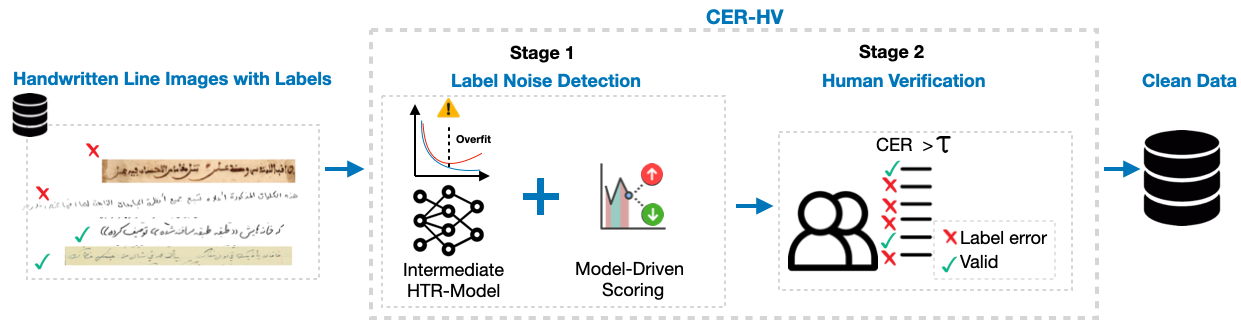}
\caption{Overview of the CER-HV framework. The first stage employs the learning dynamics of the CRNN and scores each sample using CER, and the second stage applies a human verification step to the highest ranking samples ($CER > \tau$).}
\label{fig:framework}
\end{figure}

In this section, the proposed cleaning approach is described. As illustrated in Figure~\ref{fig:framework}, the framework is divided into two main stages. The first stage is an automatic procedure that scores each sample based on the CER produced by a CRNN using the initial sample labels. The second stage is a  human-in-the-loop (HITL) procedure that verifies the highest ranking samples. The goal is to identify mislabeled samples and improve the quality of both the labels and the image content in HTR datasets. The two stages together form the CER-HV framework, which stands for CER-based Ranking with Human Verification. The following subsections present the problem formulation, the CRNN model used for recognition, and the details of the CER-HV framework.

\subsection{Problem Formulation}
\label{sec:problem-formulation}
The task of HTR is to map an image of handwritten text to a sequence of characters. Let $\mathcal{X}$ denote the space of text line images in the current dataset, where each $x \in \mathcal{X}$ is a tensor in $\mathbb{R}^{H \times W \times C}$ with height $H$, width $W$, and channels $C$. Let $\Sigma$ be the alphabet of characters, and let $\Sigma^{\ast}$ be the set of all possible sequences over this alphabet. 

For an input $x$, with ground-truth transcription $y \in \Sigma^{\ast}$, 
the goal of HTR is to predict the most likely character sequence:
\begin{equation}
y^{\ast} = \arg\max_{y \in \Sigma^{\ast}} P(y \mid x)\,.
\label{eq:htr-map}
\end{equation}

In CTC based models such as the CRNN, the network outputs a sequence of frame level probability distributions over the extended alphabet $\Sigma' = \Sigma \cup \{\text{blank}\}$. The blank symbol allows the model to handle variable length alignments by representing either no character or repeated characters. Let $T$ be the length of the output sequence produced by the encoder. Each alignment path $\pi = (\pi_{1}, \dots, \pi_{T})$ belongs to $(\Sigma')^{T}$ and collapses to a final transcription through a function $B$ that removes blanks and merges repeated characters. The probability of  $y$ is defined as the sum over all valid alignment paths:
\begin{equation}
P(y \mid x; \theta) = \sum_{\pi \in (\Sigma')^{T}, B(\pi) = y} \prod_{t=1}^{T} P(\pi_t \mid x; \theta) .
\label{eq:ctc-prob}
\end{equation}
The model parameters $\theta$ are learned by minimizing the CTC loss over the training set
\begin{equation}
\theta^{\ast} = \arg\min_{\theta} \mathcal{L}_{\mathrm{CTC}}(\theta; D)\,.
\label{eq:ctc-loss}
\end{equation}

The dataset $D = \{(x_i, y_i)\}_{i=1}^{N}$ may contain noisy labels. Let $y_i^{\ast}$ denote the unknown true label and $z_i \in \{0,1\}$ indicate whether the observed label is noisy. Noise in text recognition can come from transcription mistakes, segmentation problems, orientation issues, or irrelevant content.

The  CER-HV framework aims to identify samples that are likely to have label noise. Given the model prediction, obtained as
\begin{equation}
\hat{y}_i = \text{Decode}(f_{\theta}(x_i))
\label{eq:decode}
\end{equation}
where $f_{\theta}$ denotes the recognition model, each sample is assigned a score
\begin{equation}
c_i = \mathrm{CER}(\hat{y}_i, y_i)
\label{eq:cer-score}
\end{equation}
where CER computes the normalized edit distance between the predicted and observed sequences. 
Although high CER values often indicate potential label errors, they can also arise from correctly labeled but visually difficult samples. 
For this reason, samples with high CER values are ranked as likely to have $z_i = 1$ and are later verified in the human verification stage.
 \subsection{ HTR Model: CRNN Architecture}
\label{subsec:crnn_arch}

We instantiate the recognition model $f_{\theta}$ in Eq.~\ref{eq:decode} using a CRNN for line-level HTR. While any well-performing deep learning recognizer could be used to identify potential label errors, we employ a CRNN which
we show performs robustly across all Arabic-script datasets evaluated in this study. The CRNN is computationally efficient and has a substantially lower parameter count compared with Transformer-based models.

The CRNN model performs handwritten text line recognition using CTC \cite{graves2006connectionist}, eliminating the need for explicit character-level segmentation. Given an input image of size H$\times$W (height $\times$ width), the network interprets it as a sequence of W narrow vertical slices, each representing a small visual context of the image. The model outputs a sequence of character probability distributions of length L, typically shorter than W due to spatial downsampling. During decoding, consecutive duplicate predictions are merged and blank tokens are removed to yield the final transcription.

\begin{figure}[t]
\centering
\includegraphics[width=0.8\textwidth]{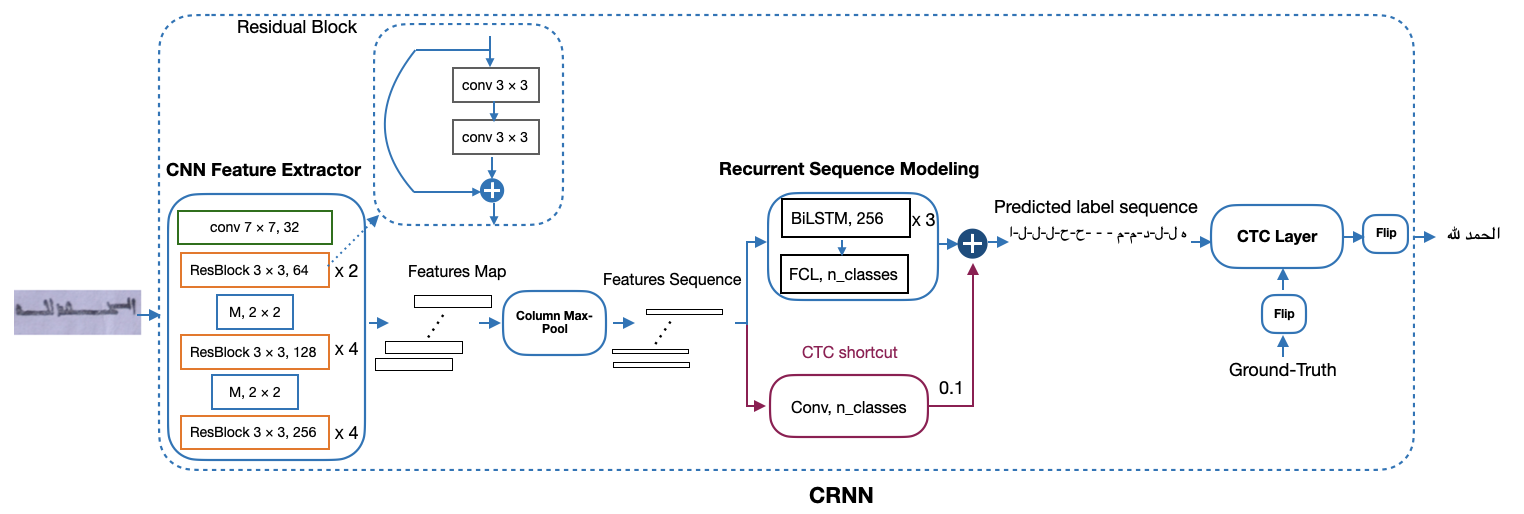}
\caption{The CRNN architecture used as the base recognition model in Stage~1 of the CER-HV framework. }
\label{fig:crnn}
\end{figure}

Our architecture builds upon the original CRNN framework proposed by Shi et al.~\cite{shi2016end}, which integrates convolutional, recurrent, and transcription components into a unified end-to-end model. We adopt the refined implementation introduced in the Best Practices framework~\cite{retsinas2022best}, illustrated in Figure~\ref{fig:crnn}. Like most CRNN-based HTR systems, the architecture combines a convolutional feature extractor, recurrent sequence modeling, and CTC decoding. However, following the Best Practices framework, our implementation incorporates three key modifications: (1) a deeper residual feature extractor with batch normalization for improved feature learning, (2) column-wise max pooling instead of concatenation to reduce feature dimensionality while maintaining vertical invariance, and (3) an auxiliary CTC shortcut branch during training to enhance recurrent layer convergence. Compared with earlier CRNN implementations based on shallow VGG-style backbones~\cite{shi2016end,puigcerver2017multidimensional}, these modifications provide stronger feature representations and improved training stability.


Concretely, the CNN feature extractor begins with a 7$\times$7 convolution (32 channels, stride 2$\times$2), followed by ResNet blocks (2 with 64 channels, 4 with 128, and 4 with 256), interleaved with batch normalization. Progressive 2$\times$2 max-pooling between block groups downsamples the spatial resolution, producing feature maps of size (H/8) $\times$ (W/8) $\times$ 256. Column-wise max pooling then flattens this tensor into a sequence of (W/8) feature vectors (256 dimensions each). 

The recurrent sequence modeling module consists of three stacked BiLSTM layers (256 hidden units each) with a dropout rate of 0.2 applied between recurrent layers, followed by an additional dropout layer with probability 0.2 before the final linear projection onto the character space. During training, CTC is applied to the resulting character logits. During inference, greedy CTC decoding selects the most probable character at each time step, merges duplicates, and removes blanks to generate the final transcription.

Following the Best Practices framework \cite{retsinas2022best}, the model employs an auxiliary CTC shortcut branch trained jointly with a weighted multi-task loss (weight = 0.1). Prior ablation experiments showed that this shortcut consistently improves CER  across multiple configurations. The shortcut branch is used only during training and is discarded during inference, introducing no additional computational overhead at test time. 

This CRNN architecture is used in the CER-HV framework as the base recognizer $f_{\theta}$. 
Its predictions $\hat{y}_i = \text{Decode}(f_{\theta}(x_i))$ are used to compute the CER scores $c_i$ that guide the ranking of samples in the noise detection stage.

\subsection{CER-HV Framework: CER Based Ranking and HITL Verification} \label{subsec:hitl}

The framework focuses on improving data quality through automated label error detection and human verification. The CRNN recognizer depends critically on the quality of the training data. Potential label noise can significantly degrade accuracy \cite{northcutt2021pervasive}. To address this, we introduce CER-HV, a two-stage framework that combines automated label noise detection based on CER scoring with human verification. Figure~\ref{fig:framework} shows the overall flow.

\textbf{Stage 1: Label Noise  Detection}

 In the first stage, we leverage the learning dynamics of deep networks (where clean samples learn early, and noisy samples fit later) for noise detection, using the CRNN architecture from Section~\ref{subsec:crnn_arch}. Here, we use the CER (a standard evaluation metric in HTR) for scoring of the samples. This represents an important difference compared to earlier approaches (e.g. O2U-Net~\cite{huang2019o2u}) built on loss-based scoring for classification tasks. In classification tasks, the training loss correlates well with label correctness, but in seq2seq problems, the CTC loss measures alignment rather than transcription accuracy. By contrast, CER provides a direct and interpretable measure of prediction quality: for example, a CER of 35\% indicates that 35\% of characters differ from the ground truth, making it immediately meaningful for human reviewers.

Another key design choice (and difference compared to earlier approaches, like O2U-Net) is that we are using an \textit{early-stopping criterion} to avoid overfitting. This approach is supported by prior studies which showed that early stopping based on validation metrics effectively detect convergence and prevent overfitting \cite{song2020does}.



We identify a single, objectively defined \textit{convergence epoch} $t_{\mathrm{conv}}$ at which training and validation CER values converge and further improvement becomes negligible. This choice simplifies the procedure, removes the need for hyperparameter tuning, and aligns better with sequence-prediction tasks such as HTR. Once the CRNN reaches the convergence epoch, each training sample is assigned a score
\begin{equation}
c_i = \mathrm{CER}(f_{\theta_{t_{\mathrm{conv}}}}(x_i), y_i)\, .
\label{eq:cer-score-single}
\end{equation}
Here, $f_{\theta_{t_{\mathrm{conv}}}}$ denotes the recognition model evaluated at the convergence epoch $t_{\mathrm{conv}}$, determined by early stopping.
The score $c_i$ measures the difference between the model prediction and the given label.
Samples are then ranked in descending order of $c_i$, with higher values indicating a greater likelihood of label noise.

\textbf{Stage 2: Human Verification}

Although automated scoring highlights suspicious samples, some correctly labeled but visually difficult examples may also receive high CER. For this reason, a HITL verification step is used. Samples whose CER exceeds a threshold are reviewed:
\[
\mathcal{S} = \{ i \mid c_i > \tau \} \qquad \tau = 0.25 \, .
\]
The threshold $\tau $ simply controls review volume and can be adjusted based on available verification budget. 
$\tau = 0.25$ was selected based on consistent observations across the five datasets that exhibited label noise, considering their training, validation, and test splits. Across these splits, samples with CER below 0.25 were overwhelmingly correctly labeled, while values above this threshold reliably indicated potential label or content errors. This value therefore represents a stable and practical operating point rather than a sensitive hyperparameter (See Section~\ref{sec:ErrorAnalysis}).  

Human reviewers inspect the selected samples and assign them to one of the following categories:
\begin{itemize}
    \item \textbf{Transcription error (TE)--} the ground truth text does not match the handwritten content in the image.
    \item \textbf{Segmentation error (SE)--} multiple text lines are included within a single image, or the lines are truncated/incomplete.
    \item \textbf{Orientation error (OE)--} significantly rotated text lines that contradict the expected reading direction.
    \item \textbf{Script mismatch (SM)--} the content is written in a script or language different from the target alphabet (e.g., Latin or numeric text in Arabic-script datasets).
    \item \textbf{Irrelevant or non-text content (IC)--} stamps, signatures or other non-textual document elements.
    \item \textbf{Valid but hard (VA)--} the sample is correct but visually challenging to correctly automatically transcribe.
\end{itemize}
Human reviewers remove or correct noisy samples while keeping the valid but hard examples. This step mitigates systematic biases in automated ranking, particularly its tendency to flag valid but challenging samples as noisy.

After human verification and cleaning, the cleaned dataset $D'$ is used to retrain the model. The retrained recognizer is evaluated on the test set, and the same CER based scoring procedure can be used to identify remaining issues in the evaluation data. The process may be repeated if needed. Algorithm~\ref{alg:hitl} summarizes the complete procedure. Beyond cleaning existing benchmarks, CER-HV can be applied as a general validation procedure to assess label quality in newly created or previously unverified line-level HTR datasets. 
\begin{algorithm}
 \footnotesize
\caption{ CER-HV Framework for Label Error Detection}

\label{alg:hitl}
\textbf{Notation:}
$D=\{(x_i,y_i)\}_{i=1}^{N}$ denotes a dataset split (training, validation, or test), where $x_i$ denotes a text line image and $y_i$ denotes its ground-truth transcription;
$f_{\theta_{t_{\mathrm{conv}}}}$ is the recognizer evaluated at the convergence epoch;
$\hat{y}_i = \mathrm{Decode}(f_{\theta_{t_{\mathrm{conv}}}}(x_i))$ is the predicted transcription;
$c_i = \mathrm{CER}(\hat{y}_i, y_i)$ is the CER score;
$\mathcal{S}$ is the set of selected high-CER samples;
$D'$ is the cleaned dataset;
$M$ is the final trained model;
$T_{\max}$ denotes the maximum number of training epochs;
$\theta$ denotes the trainable parameters of the recognizer;
$\theta_t$ denotes the parameters after epoch $t$;
$\tau$ denotes the CER threshold used to select samples for human inspection.
\vspace{0.8em}

\KwIn{Training set $D=\{(x_i,y_i)\}_{i=1}^{N}$}
\KwOut{Cleaned set $D'$, trained model $M$}

\textbf{Stage 1  : Label Noise Detection}\;
Initialize $\theta$, learning rate $\eta$, batch size $b$, early stopping patience $P$\;
\For{$t=1$ \KwTo $T_{\max}$}{
    Train the CRNN on mini batches from $D$ using the CTC loss\;
    Update parameters $\theta_t = \theta_{t-1} - \eta \nabla_{\theta}\mathcal{L}_{\mathrm{CTC}}$\;
    Compute validation CER$(t)$\;
    \If{validation CER does not improve for $P$ epochs}{
        Set convergence epoch $t_{\mathrm{conv}} = t$\;
        \textbf{break}\;
    }
}
Compute per sample CER at convergence:\;
\For{each $i$ in $D$}{
    $c_i = \mathrm{CER}(f_{\theta_{t_{\mathrm{conv}}}}(x_i), y_i)$\;
}

\textbf{Stage 2: Human Verification}\;
Rank all samples by descending $c_i$\;
Select samples $\mathcal{S} = \{i \mid c_i > \tau\}$ for human inspection\;
\For{each $(x_i,y_i) \in \mathcal{S}$}{
    Human inspects sample and determines error type:
    
    \uIf{sample has transcription, segmentation, orientation, or irrelevant content}{Fix or remove sample\;}
    \Else{Keep sample as valid but hard\;}
}
Construct the cleaned dataset $D'$ \;

Reinitialize model parameters and retrain the CRNN on $D'$ with early stopping to obtain final model $M$\

\While{further corrections may remain}{
    Optionally repeat Stages 1 to 2 on the updated $D'$\;
}

\end{algorithm}


\section{Experimental Setup }
\label{sec:experimental-setup}
This section describes the datasets used, the training details, hyperparameters, input preprocessing, and experimental settings.
\subsection{Datasets}
\label{sec:datasets}
\begin{figure}[t]
\centering

\includegraphics[width=0.7\textwidth]{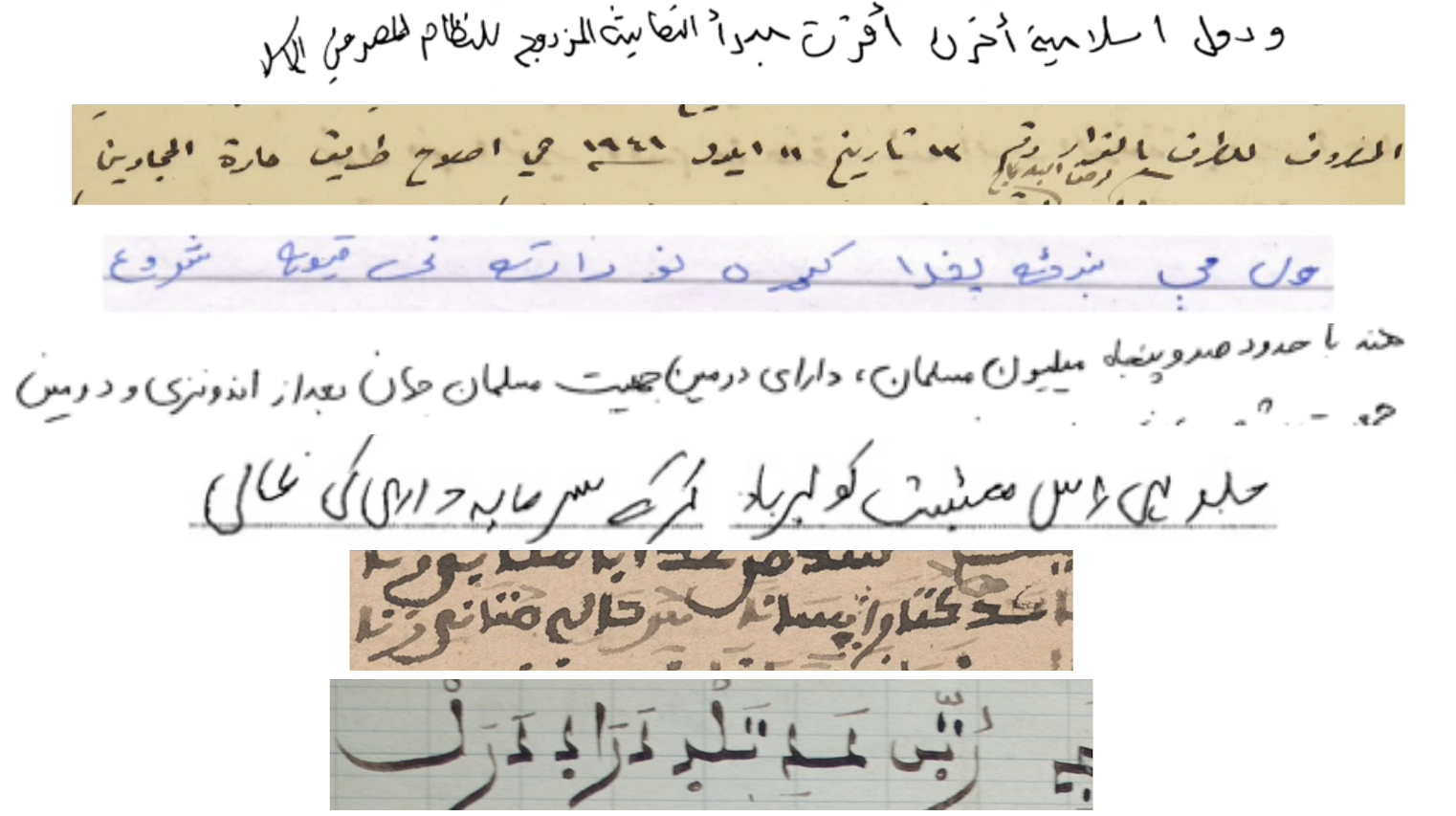}
\caption{Examples of handwriting from six datasets. From top to bottom: KHATT (Arabic), Muharaf (Arabic), PHTI (Pashto), PHTD (Persian), NUST-UHWR (Urdu), Ajami Hausa, and Ajami Fulfulde.}
\label{fig:dataset}
\end{figure}

This study employs six Arabic-script HTR datasets covering five languages segmented and labeled at the line level, covering both modern and historical handwriting. Together, these datasets provide broad coverage of writing styles, domains, and regional variations across Arabic, Persian, Urdu, Pashto, and West African Ajami scripts. Examples from the datasets are shown in Figure~\ref{fig:dataset}, and dataset statistics including train/validation/test splits are summarized in Table~\ref{tab:dataset_stats}. 
For PHTD and Ajami without public train, validation, and test splits we applied an 80\%, 10\%, and 10\% division.
These datasets collectively enable assessment of the robustness of the proposed framework across diverse handwriting styles and linguistic varieties.

\begin{table}[t]
\centering
\fontsize{8}{8}\selectfont
\begin{tabular}{l c c c c }
\hline
\textbf{Dataset} & \textbf{Language} &\textbf{Train/Val/Test}  & \textbf{Avg. W--H} & \textbf{Max Len} \\ \hline
KHATT & Arabic &4837/938/967 &  2075--200 & 132 \\ \hline
Muharaf & Arabic & 22091/1069/1334 & 575--60 & 181 \\ \hline
PHTI & Pashto & 25200/5400/5400  &  992--77  & 180 \\ \hline
PHTD & Persian &  1787 
& 1960--178  & 197 \\ \hline
NUST-UHWR & Urdu &  8483/1061/1061 & 950--64 & 116 \\ \hline
Ajami & Hausa + Fulfulde &  6132  &  1290--215 & 106 \\ \hline
\end{tabular}

\caption{Dataset descriptions including train/val/test split sizes, followed by average line image dimensions (Width--Height) and character maximum length. }
\label{tab:dataset_stats}
\end{table}

\medskip
\noindent\textit{KHATT (Arabic):} The KHATT dataset~\cite{mahmoud2014khatt} is a widely used benchmark for modern Arabic handwriting recognition. It contains paragraph images written by 1,000 writers from diverse demographic backgrounds. Each writer contributed both repeated and unique texts under controlled imaging conditions (black ink on white background). 
We used only the unique-text subset, yielding 6,742 segmented line images.

\medskip

\noindent\textit{Muharaf (Arabic):}
The Muharaf dataset~\cite{saeed2024muharaf} comprises historical Arabic 
ma\-nuscripts digitized and manually transcribed at the line level. It features considerable variation in calligraphy styles, ink intensity, and document aging effects, making it particularly challenging for modern HTR systems. Our experiments use the public release containing approximately 24K line images.

\medskip
\noindent\textit{PHTI (Pashto):}
The Pashto Handwritten Text Imagebase (PHTI) is a dataset containing 36,082 handwritten text-line images along with ground truth annotations. The dataset encompasses diverse genres including short stories, historical narratives, poetry, and religious content, collected from a variety of modern writers.  

\medskip
\noindent\textit{PHTD (Persian):}
The PHTD dataset~\cite{alaei2012dataset} consists of 140 handwritten Persian pages produced by 47 writers across three categories of content: historical documents, school dictations, and general writing. The original dataset provides pixel-level ground-truth masks for each page but does not include predefined line images or bounding boxes or train–validation–test splits.
To generate consistent and clean line images, a binary mask was generated for each line 
 from the pixel-level annotations provided with the dataset. The minimal bounding box enclosing all labeled pixels was computed with a small padding of 5~px, and the cropped region was masked to retain only the target line while setting all other pixels to background. This procedure ensures accurate isolation of individual lines despite the substantial overlap between bounding boxes caused by ascenders and descenders in cursive Persian handwriting.

We performed a similarity analysis to prevent data leakage between training and evaluation splits. This revealed that 2 pages were exact duplicates, and among the 9,730 possible page pairs, 1,155 pairs (11.9\%) exhibited more than 85\% textual similarity. Only 37 pages showed no significant overlap with any other page. The validation and test sets were selected exclusively from these 37 non-overlapping pages.
The cleaned line images and the final train, validation, and test splits will be released publicly\footnote{PHTD cleaned dataset: \url{https://huggingface.co/datasets/sana-ngu/PHTD}}.

\medskip
\noindent\textit{NUST-UHWR (Urdu):}
The NUST-UHWR dataset~\cite{ul2022convolutional} is the first published large-scale, unconstrained Urdu handwriting corpus spanning seven topical domains to ensure linguistic variety. It includes 10,608 text lines written by nearly 1,000 contributors. 

\medskip
\noindent\textit{Ajami (Hausa and Fulfulde):}
The term “Ajami” originates from Arabic and means “foreign”; it refers to African languages written using the Arabic script. The Ajami dataset ~\cite{yousuf2025handwritten} is the first publicly released benchmark for HTR of West African manuscripts, comprising 29 manually segmented and transcribed documents written in Fulfulde and Hausa.  The initial corpus contains 6,132 line images, showcasing a significant challenge for HTR systems due to its diversity of script styles, including Maghrebi, Hausa Naskh, and the distinct Hausa Thuluth calligraphy. 

The baseline Arabic HTR models used for evaluation by the dataset creators were trained on Middle Eastern and South Asian manuscripts, and yielded high CERs ranging from 65–84\% when applied to this corpus. We prepared and publicly released the extracted line images and data splits used in this work as an open-source resource to ensure reproducibility for future research.

\subsection{Training Setup and Evaluation}
\paragraph{Training setup} 
All experiments were conducted on NVIDIA A100-SXM4-40GB GPUs using a CRNN architecture configured for Arabic-script handwriting. 

For each dataset, image preprocessing followed a standardized resizing and padding procedure to reduce shape distortion and preserve aspect ratio. All line images were resized to the rounded average height and width (in whole pixels) reported in Table~\ref{tab:dataset_stats}. Images exceeding these dimensions were proportionally rescaled; otherwise, they were centered and padded with median pixel values. 

To further regularize training and enhance robustness to handwriting variability, we applied a consistent set of data augmentations across all datasets. These include affine transformations (small rotation and skew), grid distortion, elastic deformation, Gaussian noise addition, brightness and contrast adjustment, and morphological operations. 
Such augmentations have been shown to improve generalization in  HTR tasks by encouraging invariance to writer style, line curvature, and ink intensity.

For each dataset, a separate CRNN model was trained for a maximum of 800 epochs with a batch size of 16 using the Adam optimizer and an initial learning rate of $5 \times 10^{-4}$. A multi-step learning rate scheduler reduced the learning rate by a factor of 0.1 at 50\% and 75\% of the maximum number of epochs. Early stopping with patience $P = 20$ epochs was applied based on validation CER to prevent overfitting. For each dataset, the model achieving the lowest CER on the validation set was selected for evaluation on the test partition.

\paragraph{Evaluation Metric}
Recognition performance was assessed using the \textit{Character Error Rate (CER)}, the standard metric in HTR. CER is computed as the normalized Levenshtein (edit) distance between the predicted and ground-truth transcriptions, reflecting the proportion of character insertions, deletions, and substitutions required to match the prediction to the reference. Lower CER indicates higher transcription accuracy.


\section{Results}
This section presents the experimental results of our study. 
We begin by comparing our CRNN results with prior work across all datasets using the original, unaltered data to establish quantitative CER baselines. We then apply the CER-HV framework to analyze label noise across the training, validation, and test splits. Representative qualitative examples of common labeling issues are presented to illustrate the identified error categories. Finally, we quantify the impact of correcting these errors by evaluating how dataset cleaning influences recognition performance.

\subsection{Quantitative Benchmarking of the CRNN-Based HTR System}
The proposed CRNN architecture (Section \ref{subsec:crnn_arch}) was evaluated across the six Arabic-script datasets described in Section \ref{sec:datasets} on the original data without any data cleaning. Table~\ref{tab:result} summarizes the performance of this CRNN in comparison with previously reported results. For our CRNN results, we report mean CER with standard deviation across three independent training runs.

\begin{table}[!th]
\centering
\fontsize{8}{8}\selectfont
\begin{tabular}{l l c c}
\hline
\textbf{\shortstack{Dataset\\(Language)}} & \textbf{Architecture} & \textbf{Syn.} & \textbf{CER (\%)} \\
\hline

\multirow{8}{*}{\shortstack{KHATT\\(Arabic) \cite{mahmoud2014khatt}}}
 & SFR \cite{saeed2024muharaf} & Yes & 14.1 \\
 & EfficientNetB4+BLSTM \cite{bouchal2025towards} & Yes & 13.34 \\
 & FCN \cite{yousef2020accurate} & No & 8.77 \\
 & CRNN (Ours) & No & \textbf{8.46 $\pm$ 0.03} \\
 \cdashline{2-4}
 & Transformer (cross-attn.) \cite{momeni2024transformer} & Yes & 18.45 \\
 & TrOCR \cite{chan2024hatformer} & Yes & 15.4 \\
 & DAN \cite{aljishi2024comparative} & No & 8.9 \\
\hline

\multirow{3}{*}{\shortstack{Muharaf\\(Arabic) \cite{saeed2024muharaf}}}
 & SFR \cite{saeed2024muharaf} & Yes & 18.1 \\
 & CRNN (Ours) & No & \textbf{10.11 $\pm$ 0.09} \\
 \cdashline{2-4}
 & TrOCR \cite{chan2024hatformer} & Yes* & 11.7 \\
\hline

\multirow{2}{*}{\shortstack{PHTI\\(Pashto) \cite{hussain2022phti}}}
 & MD-LSTM \cite{hussain2024deep} & No & 20.7 \\
 & CRNN (Ours) & No & \textbf{8.22 $\pm$ 0.04} \\
\hline

\multirow{1}{*}{\shortstack{PHTD\\(Persian) \cite{alaei2012dataset}}}
 & CRNN (Ours) & No & \textbf{11.3 $\pm$ 0.07} \\
\hline

\multirow{8}{*}{\shortstack{NUST-UHWR\\(Urdu) \cite{ul2022convolutional}}}
 & Modified CRNN \cite{ul2022convolutional} & No & 7.35 \\
& Modified CRNN+LM \cite{ul2022convolutional} & No & \textbf{5.49} \\

 & ET-Network (No Attention) \cite{hamza2024network} & Yes & 7.79 \\
 & CRNN (Ours) & No & 6.70 $\pm$ 0.15 \\
 \cdashline{2-4}
 & ViT+T5+GPT2 \cite{hassan2025tda} & No & 17.0 \\
& Conv-Transformer \cite{riaz2022conv} & No & 6.4 \\
 & CNN-Transformer-Unified \cite{maqsood2023unified} & Yes &  \\
 & ET-Network (with Attention) \cite{hamza2024network} & Yes & 5.70 \\

\hline
\multirow{2}{*}{\shortstack{Ajami\\(Fulfulde/Hausa) \cite{yousuf2025handwritten}}}
 & OPEN-ITI\cite{yousuf2025handwritten} & No & 64--84 \\
 & CRNN (Ours) & No & \textbf{10.59 $\pm$ 0.32} \\
\hline

\end{tabular}
\caption{Comparison of our CRNN with established methods from the literature. Syn. indicates the use of synthetic or additional data. The TrOCR result on Muharaf (marked as 'Yes*' in the table) is reported as using additional 11K restricted samples in the original paper. Results are reported as CER (mean $\pm$ standard deviation). Dashed lines separate non-Transformer and Transformer-based approaches.}
\label{tab:result}
\end{table}

While most CNN–BiLSTM–CTC models share a similar high-level structure, implementation details can strongly influence recognition performance. As described in Section~\ref{subsec:crnn_arch}, our CRNN follows the Best Practices framework~\cite{retsinas2022best}, which contributes to the consistent performance gains observed across datasets.
On the \textbf{KHATT} dataset, our model achieves a new SoTA CER of \textbf{8.46\%}, outperforming both traditional CRNN-based methods~\cite{bouchal2025towards} and recent Transformer based systems such as TrOCR~\cite{chan2024hatformer} and DAN~\cite{aljishi2024comparative}, despite not using synthetic data. 

For \textbf{Muharaf}, our CRNN achieves a CER of 10.11\% using the public subset only, outperforming the baseline model~\cite{saeed2024muharaf} by eight absolute percentage points. Furthermore, our model surpasses the Transformer-based TrOCR system from~\cite{chan2024hatformer}, which reports 11.7\% CER on Muharaf using both the public and restricted portions of the dataset as well as large-scale synthetic data.

For \textbf{PHTI}, our CRNN achieves a substantial improvement over prior non-transformer models, reducing CER from 20.7\% to 8.22 \%, representing one of the largest performance gains across all evaluated datasets. For \textbf{PHTD}, where publicly reported benchmarks are limited, our model establishes a new baseline of 11.3\% CER. On \textbf{NUST-UHWR}, our model reaches 6.70\% CER, improving upon earlier convolutional and recurrent baselines~\cite{ul2022convolutional, hamza2024network}, although Transformer-based architectures~\cite{maqsood2023unified, hamza2024network} still achieve superior performance due to attention mechanisms and synthetic pretraining. 
For \textbf{Ajami}, the model attains 10.59\% CER, far outperforming previously reported results (64-84\%), which were obtained through zero-shot evaluation using pretrained OPEN-ITI models~\cite{yousuf2025handwritten}. 

In summary, the results demonstrate that a carefully optimized CRNN, trained without transformer attention or synthetic augmentation, can match or surpass more complex architectures on multiple Arabic-script datasets.

\subsection{Label Error Analysis}
\label{sec:ErrorAnalysis}
Using the CER-HV process described in Section~3.3, we flagged then analyzed label and content errors across all six datasets and across all dataset splits (training, validation, and test). The analysis reveals that the presence, type, and prevalence of labeling issues vary substantially depending on both the dataset and the split. In total, five datasets contained mislabeled or corrupted samples. \textbf{PHTD} did not exhibit any detectable labeling errors and is therefore excluded from the subsequent error analysis. 

To evaluate how accurately the proposed CER-based ranking identifies mislabeled samples, we measured the precision of the noise detector on the \textbf{top 50 highest-ranked samples} in each split. While potential errors were flagged across all datasets, precision is reported only for datasets with a sufficiently large number of flagged samples—namely \textbf{Muharaf}, \textbf{PHTI}, and \textbf{Ajami}. The remaining datasets (\textbf{KHATT} and \textbf{NUST-UHWR}) contain fewer than 50 flagged samples per split, making the top-50 precision metric not applicable. Table~\ref{tab:precision_noise_detection} reports the proportion of flagged samples that were confirmed by human reviewers to contain content errors.

The detector achieves 90\% precision on the \textbf{Muharaf} test split, and above 80\% in all three \textbf{PHTI} sets reflecting the strong correspondence between high CER and true annotation issues in this historical corpus
\textbf{Ajami} exhibits lower but still substantial precision (71\% in training, 70\% in validation, and 68\% in testing), which aligns with its visually diverse writing styles and challenging calligraphic forms. Overall, these results demonstrate that CER-based ranking effectively identifies problematic samples in noisy datasets, substantially reducing the amount of manual verification required.




\begin{table}[t]
\centering
\fontsize{8}{8}\selectfont
\begin{tabular}{l c c c}
\hline
\textbf{Dataset} & \multicolumn{3}{c}{\textbf{Precision (\%)}} \\
\cline{2-4}
 & \textbf{Train} & \textbf{Validation} & \textbf{Test} \\
\hline
Muharaf & 75 & 70 & 90 \\
PHTI    & 88 & 86 & 80 \\
Ajami   & 71 & 70 & 68 \\
\hline
\end{tabular}
\caption{Precision of the CER-based noise detection over the top 50 highest-ranked samples in each split for datasets with sufficient numbers of flagged samples. Precision reflects the proportion of flagged samples that were confirmed as true label errors; the remaining cases correspond to valid but hard samples.}
\label{tab:precision_noise_detection}
\end{table}

To further evaluate the robustness of the CER threshold $\tau$, we performed a threshold sensitivity analysis on the test splits. Since KHATT did not contain noisy samples in the test set after human verification, the analysis was restricted to Ajami, Muharaf, PHTI, and NUST-UHWR. Table~\ref{tab:tau_sensitivity_ajami} reports the full precision, recall, F1-score, and missed noisy sample statistics for the Ajami test split, together with the number of missed noisy samples for the remaining datasets. Complete results for all datasets are available in the project repository.
There is a consistent tradeoff between precision and recall as the threshold increases. Lower thresholds achieve higher recall by capturing a larger proportion of noisy samples, but also include more valid-but-difficult samples, reducing precision. 
Extremely high thresholds ($\tau=0.70$ and $\tau=0.80$) have high precision but reduce recall below 30\%.
Since CER-HV is intended as a HITL prioritization framework rather than a fully automatic filtering method, the threshold $\tau = 0.25$ balances noisy-sample coverage and review effort. 

\begin{table*}[t]
\centering
\fontsize{8}{8}\selectfont
\setlength{\tabcolsep}{5pt}

\begin{tabular}{c cccc|ccc}
\hline
\multirow{2}{*}{$\tau$} &
\multicolumn{4}{c|}{\textbf{Ajami}} &
\multicolumn{3}{c}{\textbf{Missed Noisy Samples}} \\
\cline{2-8}
& \textbf{P (\%)} &
\textbf{R (\%)} &
\textbf{F1 (\%)} &
\textbf{Missed}
& \textbf{Muharaf}
& \textbf{PHTI}
& \textbf{NUST-UHWR} \\
\hline
0.20 & 42.75 & 68.60 & 52.68 & 27 & 35 & 0 & 0 \\
0.25 & 57.58 & 66.28 & 61.62 & 29 & 39 & 0 & 0 \\
0.30 & 74.32 & 63.95 & 68.75 & 31 & 46 & 0 & 1 \\
0.35 & 78.46 & 59.30 & 67.55 & 35 & 50 & 0 & 1 \\
0.40 & 81.03 & 54.65 & 65.28 & 39 & 54 & 0 & 1 \\
0.45 & 87.04 & 54.65 & 67.14 & 39 & 60 & 4 & 2 \\
0.50 & 86.54 & 52.33 & 65.22 & 41 & 63 & 6 & 2 \\
0.70 & 89.29 & 29.07 & 43.86 & 61 & 78 & 52 & 3 \\
0.80 & 91.67 & 25.58 & 40.00 & 64 & 92 & 66 & 4 \\

\hline
\end{tabular}
\caption{Threshold sensitivity analysis. Precision, recall, F1-score, and missed noisy samples are reported for the Ajami test split. For Muharaf, PHTI, and NUST-UHWR, only the number of missed noisy samples is reported. KHATT is omitted because no noisy samples were identified in the test split during human verification. Complete threshold sensitivity results for all datasets are available on GitHub.}
\label{tab:tau_sensitivity_ajami}
\end{table*}

\begin{figure}[t]
\centering
\includegraphics[width=0.7\linewidth]{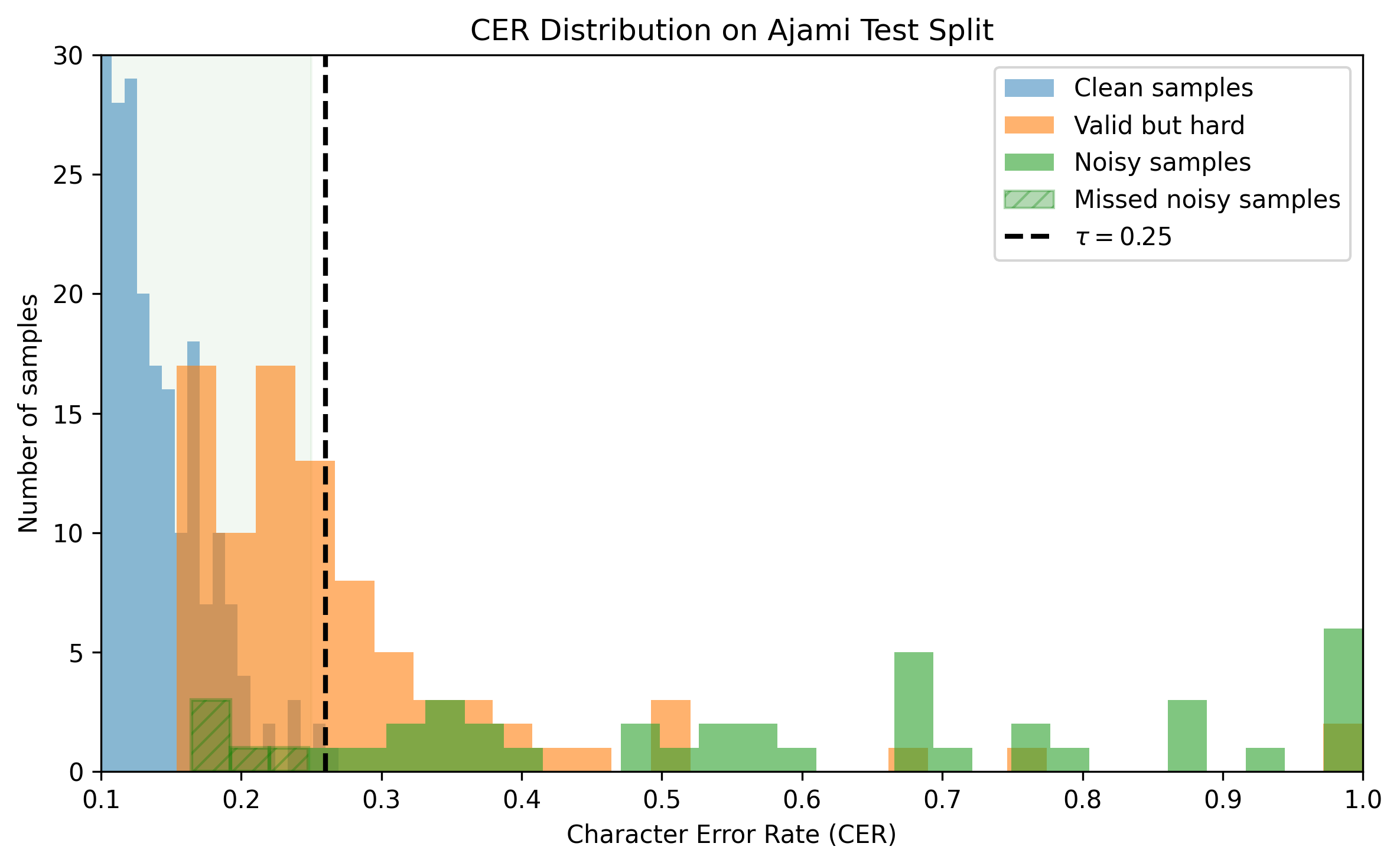}
\caption{ CER distributions for human verified noisy samples, valid but hard samples, and clean samples on the Ajami test split showing the operational threshold $\tau = 0.25$ used. Green striped bars indicate failure modes. For visualization purposes, the y-axis is truncated at 30 samples which clips the clean-sample maximum of 160 samples.}
\label{fig:ajami_cer_distribution}
\end{figure}

The CER distributions in Figure~\ref{fig:ajami_cer_distribution} illustrate that clean samples are concentrated at low CER values, whereas noisy samples accumulate at higher CER ranges. Valid but hard samples occupy an intermediate region, resulting in partial overlap with noisy samples. The CER-based ranking effectively separates clean and problematic samples with the overlap 
motivating the human verification stage in CER-HV.
At $\tau = 0.25$, only approximately 10\% of the Ajami and Muharaf test samples required inspection, while the number of missed noisy samples remained relatively low across all datasets. Cleaner datasets such as PHTI and NUST-UHWR required review of less than 5\% and 2\% of the test samples, respectively.

\begin{figure}[t]
\centering
\includegraphics[width=0.7\linewidth]{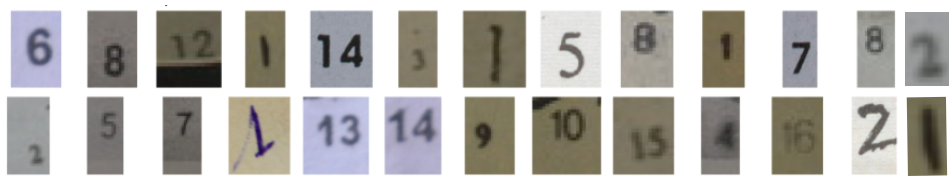}
\caption{Representative CER-HV failure modes from the Ajami test split at $\tau=0.25$.}
\label{fig:ajami_missed_samples}
\end{figure}

Figure~\ref{fig:ajami_missed_samples} shows the noisy samples that fall below the selected threshold $\tau = 0.25$ and therefore contribute to missed detections. Most missed noisy samples in both Ajami and Muharaf correspond to script mismatch cases, primarily isolated numerals
that HTR can often recognize accurately even when they originate from script-mismatch. This results in relatively low CER values, making such samples difficult to distinguish from valid samples using CER alone.

\begin{table}[t]
\centering
\fontsize{8}{8}\selectfont
\renewcommand{\arraystretch}{1.1}

\begin{tabular}{l l c c c c c c c}
\hline
{\textbf{Dataset}} &
{\textbf{Split}} &
{\textbf{TE}} &
{\textbf{SE}} &
{\textbf{OE}} &
{\textbf{SM}} &
{\textbf{IC}} &
\multicolumn{2}{c}{{\textbf{Total}}} \\
\hline

KHATT     & Train & 40 & 4  & --- & --- & --- & 44  & (0.9\%) \\
          & Val   & 5  & ---& --- & --- & --- & 5   & (0.05\%)\\
          & Test  & ---& ---& --- & --- & --- & --- & (---)  \\ \hline

Muharaf   & Train & 42 & 99 & 4   & 998 & 288 & 1431& (6.4\%)\\
          & Val   & 20 & 17 & --- & 54  & 10  & 101 & (9.4\%)\\
          & Test  & 9  & 23 & 7   & 70  & 17  & 126 & (9.4\%)\\ \hline

PHTI      & Train & 220& 2  & 98  & --- & --- & 320 & (1.3\%)\\
          & Val   & 59 & 2  & 15  & --- & --- & 76  & (1.4\%)\\
          & Test  & 48 & 4  & 20  & --- & --- & 72  & (1.3\%)\\ \hline

NUST-UHWR & Train & 3  & ---& --- & --- & --- & 3   & (0.03\%)\\
          & Val   & 4  & ---& --- & --- & --- & 4   & (0.04\%)\\
          & Test  & 5  & ---& --- & --- & --- & 5   & (0.5\%) \\ \hline

Ajami     & Train & 68 & 47 & 54  & 196 & 6   & 371 & (8.5\%)\\
          & Val   & 13 & 10 & 15  & 43  & 2   & 83  & (8.9\%)\\
          & Test  & 20 & 10 & 10  & 44  & 2   & 86  & (9.2\%)\\
\hline

\end{tabular}
\caption{Distribution of detected label errors by type across training, validation, and test splits. TE: Transcription Error; SE: Segmentation Error; OE: Orientation Error; SM: Script Mismatch; IC: Irrelevant Content.}
\label{tab:error_types_all}
\end{table}

\subsection{Categories of Flagged Samples}
\begin{figure*}[h] 
\centering
\includegraphics[width=0.8\textwidth]
{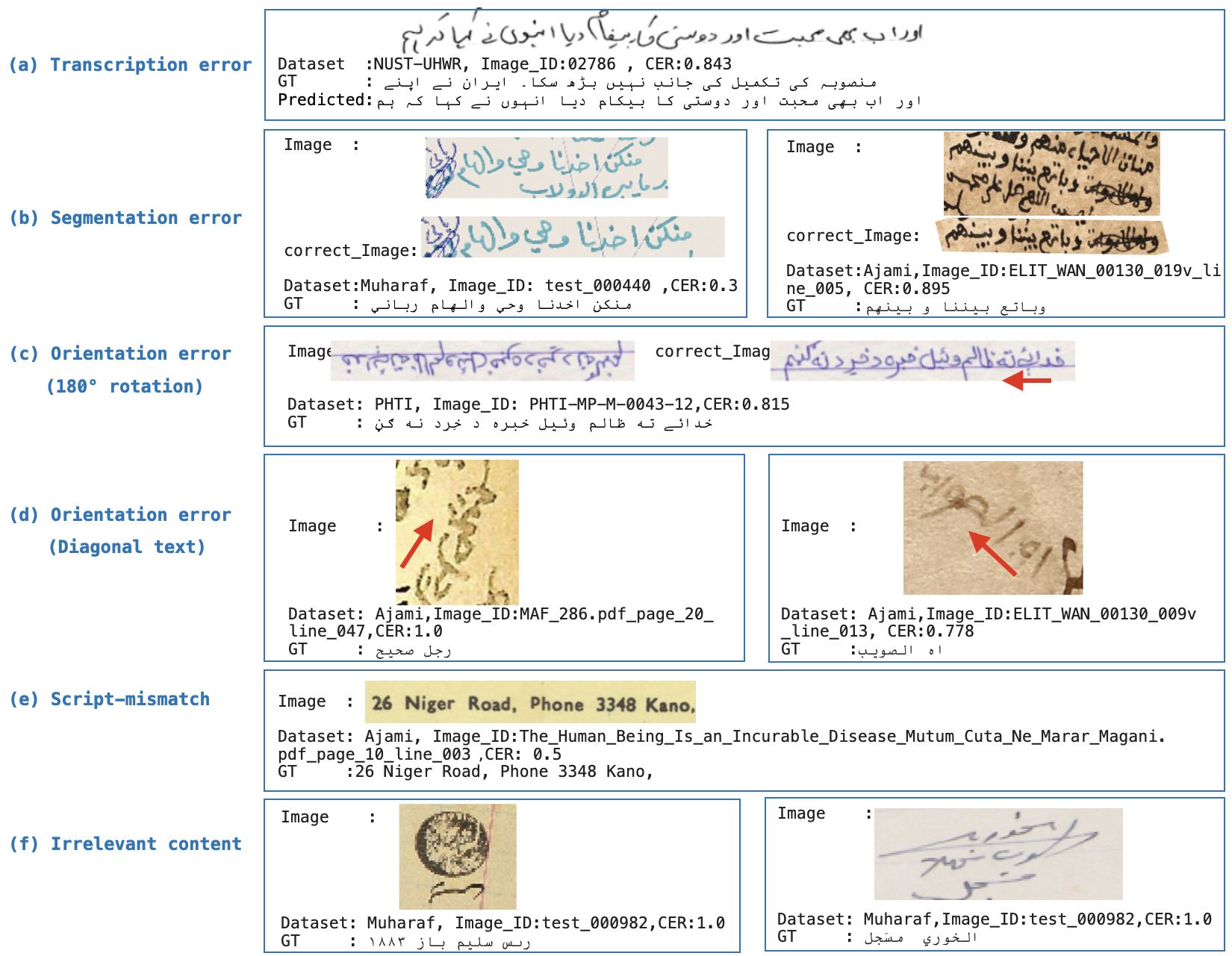}
\caption{
Representative examples of flagged samples from the major error categories identified through the CER-HV pipeline:
(a) transcription error,
(b) segmentation error,
(c) and (d) orientation error involving a 180° rotation or diagonally oriented text (red arrow indicates the reading direction),
(e) script mismatch, and
(f) irrelevant or non-text content.
For transcription errors (a), the model prediction is shown to illustrate the mismatch with the ground truth. For structural errors where correction is straightforward (b–c), the corrected image is displayed. Dataset name, image ID, CER, and ground-truth text are shown for all samples.
}
\label{fig:corrected_examples}
\end{figure*}

Figure~\ref{fig:corrected_examples} illustrates the major error categories identified by the proposed framework: transcription error, segmentation error, orientation error, script mismatch, and irrelevant or non-text content, highlighting the diversity of labeling challenges encountered across the Arabic-script datasets. While non-textual elements such as stamps or signatures and text written in non-target scripts are treated as errors in our HTR-focused setting, it is important to note that such elements may be meaningful or even desirable targets in other research contexts, such as writer identification, archival studies, or multilingual document analysis.

\begin{figure}[t]
\centering
\includegraphics[width=0.85\linewidth]{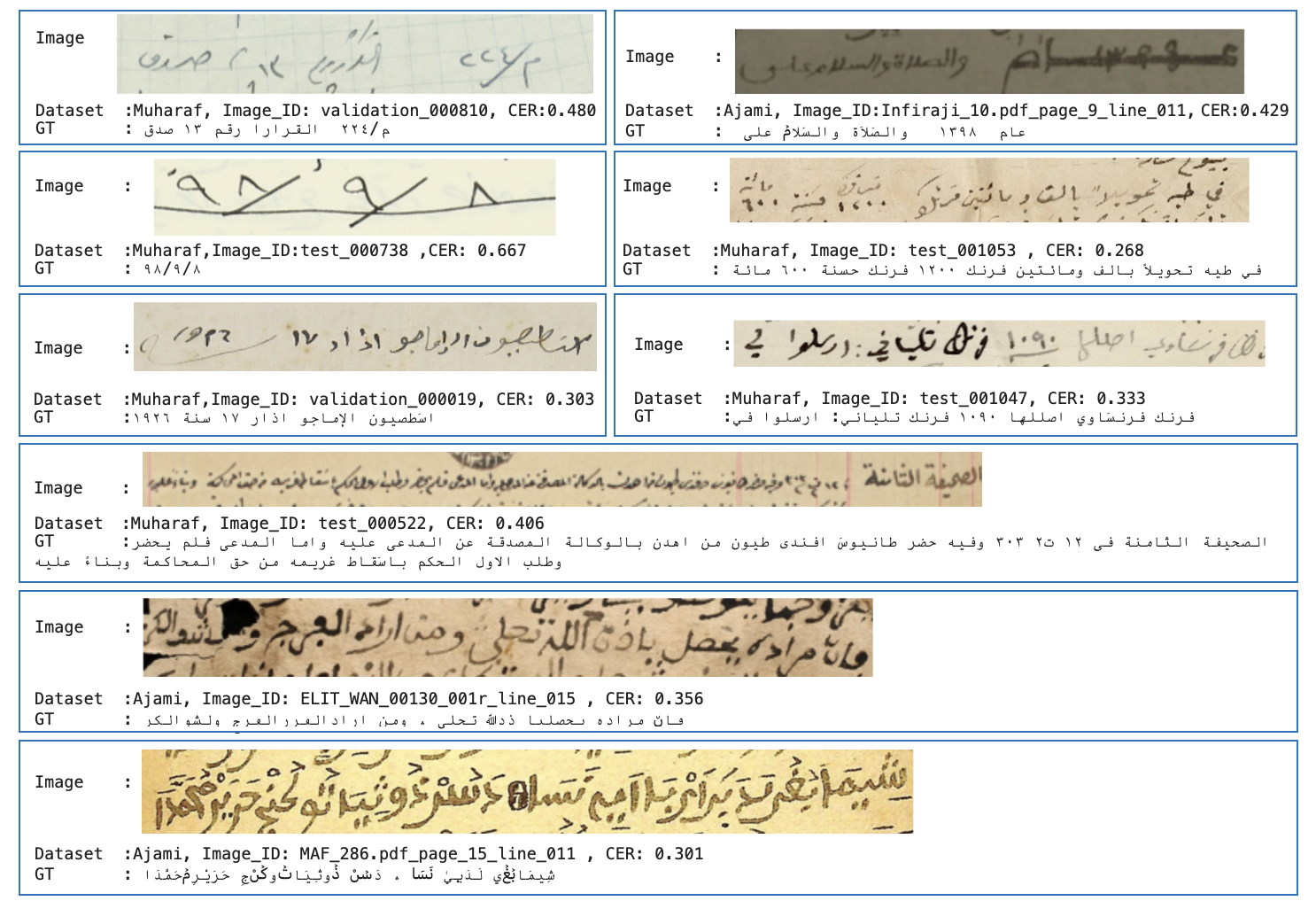}
\caption{Representative examples of ``valid but hard'' samples flagged by the detector. These cases are correctly labeled but visually challenging.}
\label{fig:valid_but_hard}
\end{figure}

In addition to noisy samples, the noise detector identified a subset of correctly labeled but visually challenging examples. These cases often involve text
containing numbers or special symbols (e.g., slashes), which are underrepresented in the training data, as well as samples with low visual clarity, unusually long text lines, heavy stroke overlap, or dense diacritics
such as shown in Figure~\ref{fig:valid_but_hard}. 
Although these samples were not corrected or removed, they provide insight into where the model still struggles and help distinguish genuine annotation errors from intrinsic recognition difficulty. 

Table~\ref{tab:error_types_all} quantifies the categories of detected error categories across the training, validation, and test splits for all datasets. Entries marked with ``---'' indicate that the corresponding error type was not observed for that dataset and split. The results show clear dataset-specific patterns. In \textbf{KHATT} and \textbf{NUST-UHWR}, errors are rare and largely limited to isolated transcription mistakes. In \textbf{PHTI}, transcription and orientation errors dominate across splits. The \textbf{Muharaf} dataset contains a higher proportion of segmentation errors and script-mismatch cases, particularly in the training data, while \textbf{Ajami} exhibits a mixture of transcription, segmentation, orientation, and script-mismatch errors across all splits. 

\subsection{Quantitative Impact of Label Errors}
Having identified the main categories of label errors, we now examine how correcting these errors affects model performance. Using the proposed CER-HV framework, we detected mislabeled or corrupted samples across all data splits and evaluated the impact of excluding these samples. We then analyzed how training on cleaned data further improves recognition performance. The combined effects of evaluation-set and training-set cleaning are summarized in Table~\ref{tab:train_clean_vs_raw_final}.

\begin{table}[!ht]
\centering
\fontsize{8}{8}\selectfont
\setlength{\tabcolsep}{4pt}

\begin{tabular}{l l l cc c ccc}
\hline
\textbf{Dataset} & \textbf{Tr. Set} & \textbf{Ev. Set} &
\multicolumn{2}{c}{\textbf{Validation Set}} & &
\multicolumn{3}{c}{\textbf{Test Set}} \\
\cline{4-5} \cline{7-9}
& & &
\textbf{CER (\%)} & \textbf{$\Delta$} & &
\textbf{CER (\%)} & \textbf{$\Delta$} & \textbf{$p$} \\
\hline

KHATT
& Raw & Raw
& 8.54 $\pm$ 0.05 & -- &&
8.46 $\pm$ 0.03 & -- & -- \\
& Raw & Clean
& 8.20 $\pm$ 0.14 & -0.34 &&
-- & -- & -- \\
& Clean & Clean
& \textbf{7.96 $\pm$ 0.05} & -0.58 &&
\textbf{8.41 $\pm$ 0.10} & -0.05 & 0.423 \\
\hline

Muharaf
& Raw & Raw
& 9.73 $\pm$ 0.09 & -- &&
10.11 $\pm$ 0.09 & -- & -- \\
& Raw & Clean
& 8.77 $\pm$ 0.03 & -0.96 &&
8.47 $\pm$ 0.10 & -1.64 & -- \\
& Clean & Clean
& \textbf{8.65 $\pm$ 0.09} & -1.08 &&
\textbf{8.26 $\pm$ 0.06} & -1.85 & \textbf{0.004} \\
\hline

PHTI
& Raw & Raw
& 8.33 $\pm$ 0.03 & -- &&
8.22 $\pm$ 0.04 & -- & -- \\
& Raw & Clean
& 7.36 $\pm$ 0.03 & -0.97 &&
7.42 $\pm$ 0.04 & -0.80 & -- \\
& Clean & Clean
& \textbf{7.26 $\pm$ 0.05} & -1.07 &&
\textbf{7.33 $\pm$ 0.06} & -0.89 & \textbf{<0.001} \\
\hline

\makecell[l]{NUST-}
& Raw & Raw
& 6.80 $\pm$ 0.10 & -- &&
6.70 $\pm$ 0.15 & -- & -- \\
UHWR& Raw & Clean
& 6.54 $\pm$ 0.05 & -0.26 &&
6.42 $\pm$ 0.03 & -0.28 & -- \\
& Clean & Clean
& \textbf{6.57 $\pm$ 0.03} & -0.23 &&
\textbf{6.45 $\pm$ 0.03} & -0.25 & 0.121 \\
\hline

Ajami
& Raw & Raw
& 10.09 $\pm$ 0.37 & -- &&
10.59 $\pm$ 0.32 & -- & -- \\
& Raw & Clean
& 9.52 $\pm$ 0.02 & -0.57 &&
9.42 $\pm$ 0.03 & -1.17 & -- \\
& Clean & Clean
& \textbf{8.95 $\pm$ 0.06} & -1.14 &&
\textbf{8.78 $\pm$ 0.16} & -1.81 & \textbf{0.023} \\
\hline

\end{tabular}

\caption{
Impact of training (Tr.) and evaluation (Ev.) set refinement on CER. Raw denotes the original dataset prior to cleaning. Results are reported as mean CER $\pm$ standard deviation over three independent training runs. The $\Delta$ column represents the cumulative improvement compared to the original raw baseline of each dataset. Statistical significance was assessed using paired t-tests on the test-set CER values, comparing  Raw/Raw and Cleaned/Cleaned over three independent runs. Statistically significant $p$ ($p<0.05$) are shown in bold.
}
\label{tab:train_clean_vs_raw_final}
\end{table}





Cleaning the evaluation splits results in noticeably reduced CER across all datasets. This improvement is relatively small (approximately 0.2--0.3~pp) for the cleanest datasets, such as \textbf{KHATT} and \textbf{NUST-UHWR}, which exhibit minimal annotation noise. Moderate gains of around 1 pp are observed for \textbf{PHTI}, while the largest improvement from evaluation-set cleaning is observed for \textbf{Muharaf} (approximately 1.6~pp), while \textbf{Ajami} improves by approximately 1.2~pp.
These cleaned evaluation splits therefore serve as revised benchmarks for the corresponding datasets, and \textbf{we release them publicly to support more reliable and reproducible evaluation in future HTR research}.

To assess how label noise affects the learned model representations, we applied the CER-HV procedure to the training splits and retrained the CRNN on the resulting cleaned datasets \(D'\). By comparing the second and third rows for each dataset in Table~\ref{tab:train_clean_vs_raw_final}, we can isolate the gain attributable specifically to training on cleaned data, beyond the effect of evaluation-set refinement.

For \textbf{KHATT} and \textbf{NUST-UHWR}, the proportion of noisy training samples is below 1\%, consisting primarily of isolated transcription errors. In these settings, retraining on cleaned data yields negligible additional improvement. This outcome is expected, as such low noise levels are unlikely to bias the learned representation and may even act as a mild regularizing factor during training.
A similar pattern is observed for \textbf{Muharaf}. Although the overall fraction of flagged training samples is higher, a substantial portion corresponds to script-mismatch cases. 


In contrast, the \textbf{Ajami} dataset provides clearer evidence of the benefit of training-side cleaning. With the highest noise density in the training split (8.5\%), Ajami exhibits a noticeable additional gain when retrained on cleaned data,
with validation CER improving from 9.52\% to 8.95\% and test CER improving from 9.42\% to 8.78\%. Unlike the other datasets, \textbf{Ajami} contains a higher proportion of structurally disruptive noise sources, such as segmentation and orientation errors, which directly affect the visual–text alignment learned by the model. Removing these samples before training therefore leads to a more consistent supervision signal and improved recognition performance.
This observation is further supported by the statistical significance analysis in Table~\ref{tab:train_clean_vs_raw_final}.


The larger impact of evaluation-set cleaning is expected because evaluation splits are substantially smaller than the corresponding training splits. Consequently, even a modest number of mislabeled evaluation samples can noticeably distort the reported CER, whereas the influence of noisy samples during training is reduced because they are mixed with a much larger number of correctly labeled examples. This observation is consistent with prior findings that label errors in benchmark test sets can disproportionately affect model evaluation and comparison \cite{northcutt2021pervasive}.
These results show that even limited amounts of label noise can substantially distort model assessment, emphasizing the need for systematic validation of evaluation sets to ensure reliable and fair benchmarking in HTR. 

\section{Conclusion and Future Work}

This work shows that data quality is a major factor in HTR for Arabic-script languages, yet it has received far less attention than model design. By combining a carefully configured CRNN with the CER-based scoring used in the CER-HV framework, we revealed a wide range of label and content issues across multiple Arabic-script datasets. Cleaning these errors improved evaluation results in every case, confirming that part of the difficulty reported in Arabic-script HTR stems from inconsistent or noisy labels rather than from script complexity alone. Importantly, CER-HV represents a task-specific reformulation of learning-dynamics-based noise detection that makes such approaches viable for CTC-based text recognition.

The results also demonstrate that a well-configured CRNN can achieve state-of-the-art performance on several datasets without synthetic data or Transformer attention. This highlights the importance of strong baselines and shows that deeper or more complex models are not always required to reach competitive accuracy. For example, on KHATT the CRNN achieves a CER of 8.46\%, matching or surpassing recent, more complex systems, while on the historical Muharaf dataset it achieves a clear improvement over the published baseline when evaluated on the public split.

The effectiveness of the human-in-the-loop component in CER-HV is closely tied to the characteristics of HTR. In CTC-based models, loss values are an indirect and often unreliable indicator of label correctness, whereas CER directly reflects transcription quality and is immediately interpretable by human reviewers. Because line-level HTR datasets typically contain thousands rather than millions of samples, restricting inspection to a small subset of high-CER samples makes targeted verification feasible without exhaustive manual review. 

Beyond its use in analyzing existing benchmarks, CER-HV can serve as a practical dataset validation tool. Applied during dataset creation, it provides a practical way to identify likely mislabeled samples.

Although this study focuses on HTR, the underlying two-stage procedure is applicable to other data-limited problems where annotation is costly and errors strongly affect evaluation outcomes. In domains such as medical imaging or symbol detection in engineering drawings, datasets are often small enough that targeted human verification is feasible.

A limitation of the present study is that cleaning is implemented through exclusion of noisy samples rather than correction. While this strategy is appropriate for analyzing the impact of label noise on benchmarking and avoids introducing additional sources of error, it does not fully exploit the information contained in the identified samples. 
In practice, different error types require different correction strategies: transcription errors could potentially be addressed through pseudo-label generation using trained HTR models, segmentation errors through dedicated segmentation approaches such as Hi-SAM, and orientation errors through automatic CER-guided rotation selection. For transcription errors, the released annotations additionally include both the original transcription and the corresponding model prediction, providing pseudo-label candidates that can support future research on semi-automated dataset correction rather than exclusion alone.  
Another important direction is reducing reliance on human verification through agreement-based filtering using complementary recognizers, such as CRNN and Transformer-based HTR models.


Overall, these findings show that progress in Arabic-script HTR depends not only on advanced models but also on reliable data, and that systematic dataset quality assessment provides a straightforward path toward stronger baselines, fairer evaluation, and more reproducible research.
\section{CRediT authorship contribution statement } 
\textbf{Sana Al-azzawi}: Conceptualization, Methodology, Software , Visualization, Data curation, Formal analysis, Writing- Original draft preparation. \textbf{Elisa Barney}: Validation, Supervision, Conceptualization,  Writing- Reviewing and Editing.  \textbf{György Kovács}:   Writing- Reviewing and Editing.  \textbf{Marcus liwicki}: Supervision,  Project administration, Writing- Reviewing and Editing.

\section{Acknowledgements} 
%
This research is financially supported by the European Regional Development Fund and the MARTINA-project (no. 20367152).

\section{Data Availability}

The datasets used in this study are publicly available








\bibliography{ref-20260602}
\bibliographystyle{elsarticle-num}

\end{document}